\documentclass{article}
\usepackage{ijcai26}
\usepackage{times}
\usepackage{soul}
\usepackage{url}
\usepackage[hidelinks]{hyperref}
\usepackage[utf8]{inputenc}
\usepackage[small]{caption}
\usepackage{graphicx}
\usepackage{amsmath}
\usepackage{amsthm}
\usepackage{booktabs}
\usepackage{algorithm}
\usepackage{algorithmic}
\usepackage[switch]{lineno}

\usepackage{amsfonts}
\usepackage{enumitem}
\usepackage{MnSymbol} 
\usepackage{bm}
\usepackage{dsfont}
\usepackage{multirow}
\usepackage{subfigure}
\usepackage{adjustbox}
\usepackage{diagbox}
\usepackage{pifont}
\usepackage{xcolor}
\usepackage{xspace}
\usepackage[normalem]{ulem}
\usepackage{colortbl}

\newcommand{\m}{HealthMamba\xspace}

\definecolor{darkgreen}{rgb}{0.0, 0.5, 0.0}
\newcommand\true{\textcolor{darkgreen}{\ding{51}}}
\definecolor{darkred}{rgb}{0.76, 0.13, 0.28}
\newcommand\false{\textcolor{darkred}{\ding{55}}}

\title{HealthMamba: An Uncertainty-aware Spatiotemporal Graph State Space Model for Effective and Reliable Healthcare Facility Visit Prediction}

\author{
Dahai Yu\textsuperscript{1},
Lin Jiang\textsuperscript{1},
Rongchao Xu\textsuperscript{1},
Guang Wang\textsuperscript{1}\thanks{Prof. Guang Wang is the corresponding author.}\\
\affiliations
\textsuperscript{1}Department of Computer Science, Florida State University\\
\emails
dahai.yu@fsu.edu, lin.jiang@fsu.edu, rxu@fsu.edu, guang@cs.fsu.edu
}

\begin{document}

\maketitle

\begin{abstract}
Healthcare facility visit prediction is essential for optimizing healthcare resource allocation and informing public health policy. 
Despite advanced machine learning methods being employed for better prediction performance, existing works usually formulate this task as a time-series forecasting problem without considering the intrinsic spatial dependencies of different types of healthcare facilities, and they also fail to provide reliable predictions under abnormal situations such as public emergencies. 
To advance existing research, we propose \m, an uncertainty-aware spatiotemporal framework for accurate and reliable healthcare facility visit prediction. 
\m comprises three key components: (i) a Unified Spatiotemporal Context Encoder that fuses heterogeneous static and dynamic information, (ii) a novel Graph State Space Model called GraphMamba for hierarchical spatiotemporal modeling, and (iii) a comprehensive uncertainty quantification module integrating three uncertainty quantification mechanisms for reliable prediction.
We evaluate \m on four large-scale real-world datasets from California, New York, Texas, and Florida. 
Results show \m achieves around 6.0\% improvement in prediction accuracy and 3.5\% improvement in uncertainty quantification over state-of-the-art baselines.
\end{abstract}
\section{Introduction}\label{introduction}
Healthcare facilities encompass a wide range of locations providing healthcare services, from small clinics and physicians' offices to urgent care centers and large hospitals. They are indispensable to population well-being for preventing and treating disease, enabling continuous care, and forming the backbone of resilient public health systems.
Consequently, predicting population visits to healthcare facilities is critical for resource allocation and policy planning~\cite{xu2026geogen,xu2026synhat,yu2026energymamba}. 

Recent studies on have made progress by using statistical methods~\cite{8766698}, deep learning methods~\cite{zhong2025peanut}, and hybrid approaches~\cite{jiang2025hcride,xu2025autostdiff,xiao2025multifrequency}. However, there are three key limitations of existing work.
First, prior studies predominantly focus on aggregated visit prediction, overlooking fine-grained healthcare facility types such as ambulatory clinics, hospitals, nursing facilities, and social assistance providers.
Second, most existing work formulates healthcare facility visit prediction as a purely time-series forecasting problem, neglecting spatial dependencies that could provide critical information for accurate prediction.
Third, most state-of-the-art approaches fail to deliver reliable predictions under public health emergencies and extreme weather events, such as the COVID-19 pandemic and hurricanes.

Hence, in this paper, we aim to develop an uncertainty-aware spatiotemporal framework for type-specific healthcare facility visit prediction that explicitly models spatial dependencies while providing reliable uncertainty estimates. However, there are two key challenges to achieving this.
The first challenge lies in the heterogeneity and sensitivity of health-related mobility. 
Our data analysis indicates that healthcare facility visits exhibit a strong correlation with population structure (e.g., age composition), spatial accessibility, and service availability. These characteristics lead to highly uneven visit patterns across regions and facility types, making it difficult for existing models to capture fine-grained spatiotemporal heterogeneity.
Second, reliable uncertainty quantification under abnormal situations remains challenging. During public emergencies such as pandemics, healthcare facility visit behaviors and demand distributions shift dramatically. Without explicit uncertainty modeling, predictions become overconfident and unreliable, potentially leading to misguided decision-making.

To address the above challenges and advance existing research, we propose \m, 
which comprises three key components: (i) a \textbf{Unified \uline{S}patio\uline{T}emporal \uline{C}ontext \uline{E}ncoder (STCE)}, which fuses heterogeneous static and time-varying contextual information and historical visit data into compact node-time representations, (ii) a novel \textbf{GraphMamba Backbone (G-Mamba)}, which innovatively integrates adaptive graph learning into a UNet-style Mamba architecture for hierarchical spatiotemporal modeling, and (iii) a \textbf{comprehensive uncertainty 
quantification (UQ)} module integrating node-based, distribution-based, and parameter-based UQ mechanisms, followed by a post-hoc quantile calibration to achieve higher prediction reliability.

The key contributions of this paper are as follows: 
\begin{itemize}
    \item \textbf{Conceptually}, unlike existing works that focus on aggregated healthcare facility visit prediction and treat this task as a time-series forecasting problem, we target type-specific healthcare facility visit prediction and formulate it as an uncertainty-aware spatiotemporal forecasting task with explicit spatial modeling.
    
    \item \textbf{Technically}, we propose \m, an uncertainty-aware spatiotemporal framework comprising three novel components: (i) a Unified Spatiotemporal Context Encoder for heterogeneous information fusion, (ii) a graph state space model called GraphMamba for hierarchical spatiotemporal modeling, and (iii) a comprehensive UQ module that integrates three different UQ mechanisms and a quantile calibration for reliable prediction. 
    
    \item \textbf{Empirically}, we evaluate \m on four real-world datasets from California, New York, Texas, and Florida by comparing it with 13 state-of-the-art baselines across five metrics. Extensive results demonstrate that \m achieves 6.0\% higher prediction accuracy and 3.5\% better uncertainty quantification than the best baseline. Our implementation is available at 
\url{https://github.com/UFOdestiny/HealthMamba}. A longer version of this paper can be found at \url{https://arxiv.org/pdf/2602.05286}.
\end{itemize}

\begin{figure*}[htpb]
    \centering
\includegraphics[width=\linewidth]{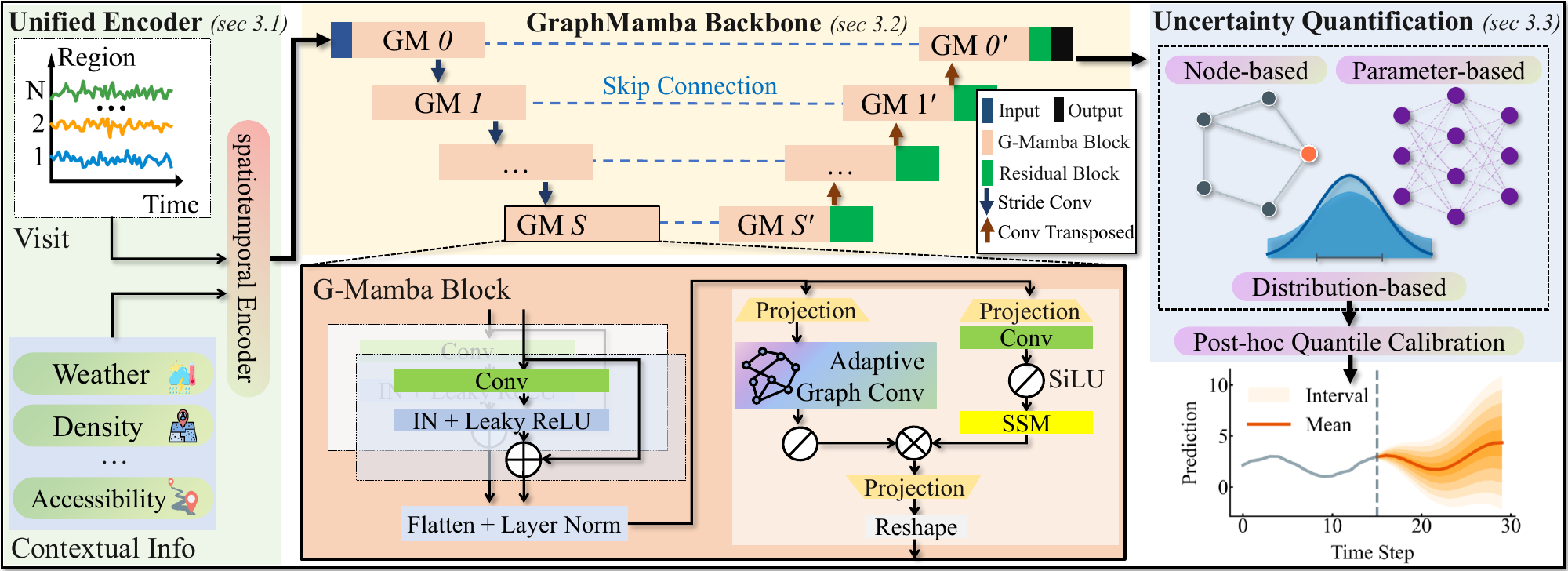}
  \caption{Framework of \m, which consists of three major components: (i) Unified Spatiotemporal Context Encoder for heterogeneous input fusion, (ii) GraphMamba Backbone for hierarchical spatiotemporal modeling with adaptive graph learning, and (iii) comprehensive uncertainty quantification (UQ) module with three types of UQ mechanisms and post-hoc quantile calibration for reliable prediction.}
  \label{fig:framework}
\end{figure*}

\section{Problem Formulation}\label{sec:problem-formulation}
Based on data analysis using real-world data, we found that there is a disparity in healthcare facility distribution and accessibility among different regions. For example, there are fewer healthcare facilities in rural areas, and rural residents need to travel longer distances to visit healthcare facilities. In addition, the visit patterns of different types of healthcare facilities are also different. Hence, we intuitively build graphs to model spatial dependencies to predict multi-type healthcare facility visits.

\subsection{Spatiotemporal Graph Construction}\label{subsec:graph-construction}
We define a spatiotemporal graph $\mathcal{G}=(\mathcal{V},\mathcal{E},\mathbf{A})$, where $\mathcal{V}=\{v_1,\ldots,v_N\}$ denotes $N$ nodes (e.g., counties), $\mathcal{E}$ is the edge set, and $\mathbf{A}\in\mathbb{R}^{N\times N}$ is the weighted adjacency matrix. At each time step $t$, node $i$ has a feature vector $\mathbf{v}_{i,t}\in\mathbb{R}^{C}$ representing visit counts across $C$ healthcare facility categories.
We construct $\mathbf{A}$ using a Gaussian kernel with sparsity thresholding:
\begin{equation}
a_{ij}=
\begin{cases}
\exp\!\left(-d_{ij}^{2}/\sigma^{2}\right), & i\neq j~\text{and}~\exp\!\left(-d_{ij}^{2}/\sigma^{2}\right)\ge \epsilon,\\[4pt]
0, & \text{otherwise},
\end{cases}
\label{eq:adjacency}
\end{equation}
where $d_{ij}$ is the centroid distance between nodes $v_i$ and $v_j$, $\sigma>0$ controls spatial decay, and $\epsilon\in[0,1]$ is a sparsity threshold that removes weak connections.

\subsection{Healthcare Facility Visit Prediction}\label{subsec:prediction}
Let $\mathbf{V}\in\mathbb{R}^{N\times T_{\mathrm{in}}\times C}$ denote the input visit tensor including features of all $N$ nodes over $T_{\mathrm{in}}$ historical time steps across $C$ healthcare facility types. 
In addition to historical visits, we also incorporate static attributes $\mathbf{D}\in\mathbb{R}^{N\times d_{\mathrm{dem}}}$ (e.g., demographics) and time-varying external factors $\mathbf{E}\in\mathbb{R}^{N\times T_{\mathrm{in}}\times d_{\mathrm{ext}}}$ (e.g., weather, accessibility) as auxiliary inputs, where $d_{\mathrm{dem}}$ and $d_{\mathrm{ext}}$ denote the dimensions of demographic attributes and external factors, respectively.
The goal of healthcare facility visit prediction is to predict future visits for the next $T_{\mathrm{out}}$ steps given historical observations and auxiliary information.

\subsubsection{Deterministic Prediction}\label{subsubsec:deterministic}
Conventional deterministic prediction learns a mapping $f$ from inputs to point forecasts:
\begin{equation}
\{\mathbf{V}, \mathbf{D}, \mathbf{E}, \mathbf{A}\}\xrightarrow{~f~}\hat{\mathbf{Y}}\in\mathbb{R}^{N\times T_{\mathrm{out}}\times C}.
\label{eq:deterministic}
\end{equation}

\subsubsection{Probabilistic Prediction}\label{subsubsec:probabilistic}
In this work, we focus on probabilistic prediction, which quantifies uncertainty using prediction intervals. Given a target miscoverage rate $\alpha$ (e.g., $\alpha=0.1$ for 90\% coverage), we predict lower, upper, and median quantiles:
\begin{equation}
\{\mathbf{V}, \mathbf{D}, \mathbf{E}, \mathbf{A}\}\xrightarrow{~\mathcal{F}~}
\big[\hat{\mathbf{L}},\,\hat{\mathbf{U}},\,\hat{\mathbf{M}}\big],
\label{eq:probabilistic}
\end{equation}
where $\hat{\mathbf{L}}=\hat{q}_{\alpha/2}(\mathbf{V})$, $\hat{\mathbf{U}}=\hat{q}_{1-\alpha/2}(\mathbf{V})$, and $\hat{\mathbf{M}}=\hat{q}_{0.5}(\mathbf{V})$ represent the lower bound, upper bound, and median predictions, respectively, all in $\mathbb{R}^{N\times T_{\mathrm{out}}\times C}$. The median $\hat{\mathbf{M}}$ also serves as the point prediction, analogous to $\hat{\mathbf{Y}}$ in Eq.~\eqref{eq:deterministic}.

\section{Methodology}\label{methodology}
In this paper, we propose \m, a spatiotemporal framework for effective and reliable healthcare facility visit prediction. An overview of \m is shown in Figure~\ref{fig:framework}, which comprises three key components: (i) a \textbf{Unified \uline{S}patio\uline{T}emporal \uline{C}ontext \uline{E}ncoder (STCE)}, which fuses heterogeneous contextual information and historical visit data into compact node-time representations, (ii) a novel \textbf{GraphMamba Backbone (G-Mamba)}, which innovatively integrates adaptive graph learning into a UNet-style Mamba architecture for hierarchical spatiotemporal modeling, and (iii) a \textbf{comprehensive uncertainty 
quantification} module integrating node-based, distribution-based, and parameter-based uncertainty quantification, followed by a post-hoc quantile calibration to achieve higher prediction reliability.

\subsection{Unified Spatiotemporal Context Encoder}
\label{STCE}

Given the input visit tensor $\mathbf{V}$, static attributes $\mathbf{D}$, dynamic external factors $\mathbf{E}$, and the adjacency matrix $\mathbf{A}$ as defined in Section~\ref{sec:problem-formulation}, 
STCE produces a unified node-time representation $\mathbf{R}\in\mathbb{R}^{N\times T_{\mathrm{in}}\times d_{\mathrm{model}}}$ that fuses these heterogeneous inputs while considering spatial structure and temporal dynamics:
\begin{align}
    \mathbf{R} = \mathrm{STCE}(\mathbf{V},\mathbf{D},\mathbf{E},\mathbf{A}).
\end{align}
We utilize $d_{\mathrm{hid}}$ and $d_{\mathrm{model}}$ to denote the hidden-layer dimension and model output embedding dimension, respectively.
Specifically, we first map inputs to a common hidden space via feature embedding layers:
\begin{align}
    \mathbf{H}^v_{i,t} &= \phi_v\big(W_v \,\mathbf{v}_{i,t} + b_v\big) \in\mathbb{R}^{d_{\mathrm{hid}}}, \\
    \mathbf{H}^d_{i} &= \phi_d\big(W_d \,\mathbf{D}_{i} + b_d\big) \in\mathbb{R}^{d_{\mathrm{hid}}}, \\
    \mathbf{H}^e_{i,t} &= \phi_e\big(W_e \,\mathbf{E}_{i,t} + b_e\big) \in\mathbb{R}^{d_{\mathrm{hid}}},
\end{align}
where \(\phi_\ast\) denotes a nonlinear function (e.g., SiLU/GELU) with dropout. The initial fused embedding at each \((i,t)\) is:
\begin{equation}
    \mathbf{X}_{i,t} = \mathbf{H}^v_{i,t} + \mathbf{H}^d_i + \mathbf{H}^e_{i,t}.
    \label{eq:init-fuse}
\end{equation}
For spatial encoding, we apply graph convolution with activation function ReLU on node features \(\mathbf{X}_{:,t}\) at each time \(t\):
\begin{equation}
    \widetilde{\mathbf{X}}_{:,t} = \mathrm{GConv}\big(\mathbf{X}_{:,t}; \mathbf{A}\big)=\sigma\!\Big(\sum_{j=1}^N \tilde{a}_{ij} \,W_g\,\mathbf{X}_{j,t} + b_g\Big).
\end{equation}
For temporal encoding, we mix information across the $T_{\mathrm{in}}$ steps for each node \(i\) using a light-weight temporal mixer combining depthwise 1D convolution and channel mixing:
\begin{align}
    \mathbf{U}_{i,:} &= \mathrm{DepthConv1D}\big(\widetilde{\mathbf{X}}_{i,:}\big) + \widetilde{\mathbf{X}}_{i,:} \in\mathbb{R}^{T_{\mathrm{in}}\times d_{\mathrm{hid}}},\\
    \widehat{\mathbf{X}}_{i,t} &= \mathrm{MLP_{chan}}\big(\mathrm{LayerNorm}(\mathbf{U}_{i,t})\big) + \mathbf{U}_{i,t},
\end{align}
where \(\mathrm{DepthConv1D}\) mixes information across the temporal axis and \(\mathrm{MLP_{chan}}\) performs channel mixing. LayerNorm and residual connections stabilize training. The final projection is:
\begin{equation}
    \mathbf{Z}_{i,t} = \mathrm{LayerNorm}\Big( W_p \, \big[ \widehat{\mathbf{X}}_{i,t} \,;\, \mathbf{H}^d_i \big] + b_p \Big),
\end{equation}
\begin{equation}
    \mathbf{R}_{i,t} = \mathrm{Dropout}\big( \mathrm{SiLU}(W_o \, \mathbf{Z}_{i,t} + b_o) \big) \in\mathbb{R}^{d_{\mathrm{model}}}.
\end{equation}
The encoder output $\mathbf{R}\in\mathbb{R}^{N\times T_{\mathrm{in}}\times d_{\mathrm{model}}}$ serves as the input to the GraphMamba backbone.

\subsection{GraphMamba Backbone}
In this part, we innovatively integrate an adaptive graph learning module into a UNet-style Mamba architecture for hierarchical spatiotemporal modeling. 
The input feature at scale $s$ is $\mathbf{X}^{(s)}\in\mathbb{R}^{N\times T_{\mathrm{in}}\times d_s}$, with $\mathbf{X}^{(0)}=\mathbf{R}$. 
The encoder applies a sequence of G-Mamba blocks followed by downsampling; 
the decoder performs symmetric upsampling followed by additional G-Mamba blocks. 
At the coarsest scale, we initialize the decoder by setting \(\mathbf{Z}^{(S)}=\mathbf{Y}^{(S)}\). For other scales, we use the following equations.

\begin{align}
    \mathbf{Y}^{(s)} &= \text{G-Mamba}^{(s)}\big(\mathbf{X}^{(s)}\big), \\
    \mathbf{X}^{(s+1)} &= \mathrm{DownConv}\big(\mathbf{Y}^{(s)}\big), \qquad s=0,\dots,S-1,\\
    \widehat{\mathbf{Y}}^{(s)} &= \mathrm{UpConv}\big(\mathbf{Z}^{(s+1)}\big), \qquad s=S-1,\dots,0,\\
    \mathbf{Z}^{(s)} &= \mathrm{Fuse}\big(\mathbf{Y}^{(s)}, \widehat{\mathbf{Y}}^{(s)}\big).
\end{align}

Here, \(\mathrm{DownConv}\) denotes stride convolution, \(\mathrm{UpConv}\) denotes transposed convolution, and \(\mathrm{Fuse}\) represents skip-connection fusion. 
Each scale maintains its own G-Mamba block without shared weights, allowing scale-specific modeling capacity. 

\subsubsection{G-Mamba Block}
Each GraphMamba Block (G-Mamba) is designed to capture spatial dependencies, temporal dynamics, and channel interactions, incorporating an Adaptive Graph Learning mechanism that learns scale-specific, data-driven graph structures.

\textbf{Adaptive Graph Learning.}
We first derive a node-level embedding by aggregating temporal information:
\begin{equation}
    \mathbf{u}_i = \mathrm{Pooling}\big(\mathbf{X}^{(s)}_{i,:}\big).
\end{equation}
Node affinities are learned using an attention-style mechanism:
\begin{align}
    \tilde{e}_{ij} &= \mathrm{LeakyReLU}\big(\mathbf{a}^\top [W_u\mathbf{u}_i \,||\, W_u\mathbf{u}_j]\big),\\
    \alpha_{ij} &= \frac{\exp(\tilde{e}_{ij})}{\sum_{k=1}^N\exp(\tilde{e}_{ik})}\in \mathbf{A}^{(\mathrm{ag})}.
\end{align}
The resulting adjacency is symmetrized and normalized:
{\small\begin{equation}
    \mathbf{A}^{(\mathrm{sym})} = \tfrac{1}{2}\big(\mathbf{A}^{(\mathrm{ag})} + (\mathbf{A}^{(\mathrm{ag})})^\top\big),\quad
    \widehat{\mathbf{A}} = \tilde{\mathbf{D}}^{-\frac{1}{2}}\mathbf{A}^{(\mathrm{sym})}\tilde{\mathbf{D}}^{-\frac{1}{2}},
\end{equation}}
where $\tilde{\mathbf{D}}$ is the degree matrix of $\mathbf{A}^{(\mathrm{sym})}$.
If prior adjacency \(\mathbf{A}_0\) is available, we blend the two sources:
\begin{equation}
    \mathbf{A}^{\ast} = \lambda\mathbf{A}_0 + (1-\lambda)\widehat{\mathbf{A}}, \quad \lambda\in[0,1].
\end{equation}

\textbf{Graph-enhanced Spatial Mixing.}
Using the learned adjacency \(\mathbf{A}^{\ast}\), the block performs graph convolution at each time step:
\begin{equation}
    \mathbf{G}_{:,t} = \sigma\!\Big(\mathbf{A}^{\ast}\,(W_g \mathbf{X}^{(s)}_{:,t}) + b_g\Big).
\end{equation}
A residual connection stabilizes learning:
\begin{equation}
    \mathbf{G} \leftarrow \mathbf{G} + \mathbf{X}^{(s)}.
\end{equation}

\textbf{Temporal and Channel Mixing.}
Temporal dependencies are modeled via a State-Space Module (SSM), followed by channel mixing:
\begin{align}
    \mathbf{T}_{i,:} &= \mathrm{SSM}\big(\mathbf{G}_{i,:}\big) + \mathbf{G}_{i,:},\\
    \mathbf{C}_{i,t} &= \mathrm{MLP_{chan}}\big(\mathrm{LayerNorm}(\mathbf{T}_{i,t})\big) + \mathbf{T}_{i,t}.
\end{align}

\textbf{Projection and Residual Refinement.}
The block output is produced by projecting back to the original dimension with a residual connection:
\begin{equation}
    \mathbf{O} = \mathrm{LayerNorm}\big(W_o \mathbf{C} + b_o\big) + \mathbf{X}^{(s)}.
\end{equation}

\textbf{Multi-scale Fusion and Skip Connections.}
During decoding, the upsampled representation at scale \(s\) is fused with the encoder's corresponding feature \(\mathbf{Y}^{(s)}\) via concatenation and projection:
\begin{equation}
    \mathbf{Z}^{(s)}=\mathrm{Fuse}(\mathbf{Y}^{(s)}, \widehat{\mathbf{Y}}^{(s)}) =
    \mathrm{Proj}\big([\mathbf{Y}^{(s)};\widehat{\mathbf{Y}}^{(s)}]\big).
\end{equation}
The fused tensor is passed through another G-Mamba block, enabling joint refinement of encoder and decoder information.

\subsection{Uncertainty-aware Prediction}

We design a modular uncertainty quantification component atop the GraphMamba backbone that produces calibrated uncertainty estimates. This component comprises three complementary mechanisms, node-based, distribution-based, and parameter-based uncertainty quantification, which together capture local heteroscedasticity, predictive distributional shape, and epistemic uncertainty.

\subsubsection{Node-based Uncertainty}
Node-based uncertainty produces per-node prediction intervals expressing local uncertainty at the region level. For a prediction target \(y_{i,t+\tau}\) at node \(i\) and horizon \(\tau\), we learn lower and upper quantile heads
$\ell_{\alpha}(i,t,\tau)$ and $u_{\alpha}(i,t,\tau)$
for a nominal miscoverage level \(\alpha\) (e.g., \(\alpha=0.1\) for a 90\% interval). Each quantile head is implemented as a projection from the backbone's final hierarchical representation \(\mathbf{Z}\). The pinball loss for a set \(\mathcal{Q}\) of quantile levels is:
\begin{align}
\mathcal{L}_{\mathrm{quant}} \;=\; \frac{1}{|\mathcal{Q}|}\sum_{q\in\mathcal{Q}} \frac{1}{|\mathcal{S}|}\sum_{(i,t+\tau)\in\mathcal{S}} \rho_q\big(y_{i,t+\tau} - \hat{q}_{i,t+\tau}\big),
\end{align}
where \(\rho_q(z)=\max(qz,(q-1)z)\) is the pinball loss, \(\hat{q}_{i,t+\tau}\) denotes the predicted \(q\)-quantile, and \(\mathcal{S}\) is the training set. Node-based intervals are robust to misspecification of the full predictive distribution, as they directly target coverage without assuming any parametric form. 

\subsubsection{Distribution-based Uncertainty}
Distribution-based uncertainty fits a parametric predictive distribution (e.g., Gaussian or Laplace) per node-time. The most common instantiation is a heteroscedastic Gaussian predictor outputting mean \(\mu_{i,t+\tau}\) and variance \(\sigma^2_{i,t+\tau}\):

\begin{align}
p(y_{i,t+\tau}\mid\mathbf{x}) = \mathcal{N}\big(y_{i,t+\tau}\,;\,\mu_{i,t+\tau},\,\sigma^2_{i,t+\tau}\big).
\end{align}
This head is trained by minimizing the negative log likelihood (NLL):
\begin{align}
\mathcal{L}_{\mathrm{nll}} \;=\; \frac{1}{|\mathcal{S}|}\sum_{(i,t+\tau)\in\mathcal{S}}
\left( \tfrac{1}{2}\log\sigma^2_{i,t+\tau} + \tfrac{(y_{i,t+\tau}-\mu_{i,t+\tau})^2}{2\sigma^2_{i,t+\tau}} \right).
\end{align}
Distribution-based heads capture aleatoric uncertainty (data noise and heteroscedasticity) and allow analytic computation of prediction intervals, e.g., \(\mu\pm z_{1-\alpha/2}\sigma\) for Gaussian.

\subsubsection{Parameter-based Uncertainty}
Parameter-based uncertainty estimates epistemic uncertainty arising from model parameters. We adopt the commonly-used MC Dropout, which performs \(M\) stochastic forward passes at inference with dropout enabled. Given \(M\) stochastic predictions \(\{\mu^{(m)}_{i,t+\tau}\}_{m=1}^M\) and \(\{\sigma^{2,(m)}_{i,t+\tau}\}_{m=1}^M\), the predictive mean and total variance decompose as:
{\small\begin{align}
\widehat{\mu}_{i,t+\tau} &= \frac{1}{M}\sum_{m=1}^M \mu^{(m)}_{i,t+\tau}, \\
\mathrm{Var}_{\mathrm{pred}}(y) &= 
\underbrace{\frac{1}{M}\sum_{m=1}^M \sigma^{2,(m)}_{i,t+\tau}}_{\text{aleatoric}} \;+\;
\underbrace{\frac{1}{M}\sum_{m=1}^M\big(\mu^{(m)}_{i,t+\tau}-\widehat{\mu}_{i,t+\tau}\big)^2}_{\text{epistemic}}.
\end{align}}
This decomposition yields interval estimates that combine both sources of aleatoric and epistemic uncertainty. The parameter-based loss minimizes the variance of ensemble predictions to encourage consistency, i.e., the model should produce similar predictions across different dropout masks when it is confident about a given input:
\begin{align}
\mathcal{L}_{\mathrm{param}} \;=\; \frac{1}{|\mathcal{S}|}\sum_{(i,t+\tau)\in\mathcal{S}} \frac{1}{M}\sum_{m=1}^M\big(\mu^{(m)}_{i,t+\tau}-\widehat{\mu}_{i,t+\tau}\big)^2.
\end{align}
A lower value of \(\mathcal{L}_{\mathrm{param}}\) indicates that the stochastic predictions are tightly clustered around their mean, reflecting lower epistemic uncertainty.

\subsubsection{Integrated Training Objective}
We jointly train all the above uncertainty heads using:
\begin{align}
\mathcal{L}_{\mathrm{total}} \;=\; \mathcal{L}_{\mathrm{quant}}
\;+\; \mathcal{L}_{\mathrm{nll}}
\;+\; \mathcal{L}_{\mathrm{param}}
\;+\; \mathcal{L}_{\mathrm{calib}},
\end{align}
where \(\mathcal{L}_{\mathrm{quant}}\), \(\mathcal{L}_{\mathrm{nll}}\), and \(\mathcal{L}_{\mathrm{param}}\) correspond to node-based, distribution-based, and parameter-based uncertainty losses, respectively. The calibration loss \(\mathcal{L}_{\mathrm{calib}}\) encourages standardized residuals to match a standard normal distribution. Over a training minibatch \(\mathcal{B}\), we compute
{\small\begin{align} 
\mathcal{L}_{\mathrm{calib}} \,=\,
\Big(\tfrac{1}{|\mathcal{B}|}\sum_{(i,t+\tau)\in\mathcal{B}} r_{i,t+\tau}\Big)^{\!2}
\;+
\Big(\tfrac{1}{|\mathcal{B}|}\sum_{(i,t+\tau)\in\mathcal{B}} r_{i,t+\tau}^{2} - 1\Big)^{\!2},
\end{align}}
which enforces the standardized residuals to have zero mean and unit variance, thereby improving probabilistic calibration. Here, the standardized residual is defined as
\begin{align}
r_{i,t+\tau} \,=\, \frac{y_{i,t+\tau}-\mu_{i,t+\tau}}{\sigma_{i,t+\tau}+\varepsilon},
\end{align}
where \(\varepsilon>0\) is a small constant for numerical stability.

\subsubsection{Post-hoc Quantile Calibration}
To improve the empirical coverage of the trained quantile intervals, we apply a post-hoc quantile calibration on a held-out calibration set. Let \((\ell_i,u_i)\) be the predicted interval and \(y_i\) be the ground truth, we compute the empirical coverage gap:
\begin{align}
\Delta_{\mathrm{cov}} = (1-\alpha) - \frac{1}{|\mathcal{C}|}\sum_{i\in\mathcal{C}} \mathbf{1}[\ell_i \leq y_i \leq u_i],
\end{align}
where \(\mathcal{C}\) is the calibration set and \(\alpha\) is the target miscoverage level. We then compute the adjustment margin:
\begin{align}
c = \mathrm{Quantile}_{1-\alpha}\big\{\max\{\ell_i - y_i,\, y_i - u_i,\, 0\}\big\}_{i\in\mathcal{C}}.
\end{align}
The calibrated interval for a new prediction becomes
\begin{align}
\big[\ell^{\mathrm{cal}}, u^{\mathrm{cal}}\big] \;=\; \big[\ell - c,\, u + c\big],
\end{align}
which adjusts the quantile regression intervals to achieve the target coverage level.
\section{Evaluation}\label{evaluation}

In this section, we conduct a comprehensive experimental evaluation of our proposed \m. Specifically, we aim to address the following five research questions:
\begin{itemize}[leftmargin=5.0mm]
\item \textbf{RQ 1}: How does \m perform compared to state-of-the-art baselines?
\item \textbf{RQ 2}: Is \m effective for visit prediction of different types of healthcare facilities?
\item \textbf{RQ 3}: Are all components in \m effective?
\item \textbf{RQ 4}: How does \m perform under abnormal scenarios such as public emergencies (COVID-19)?
\end{itemize}

\subsection{Evaluation Setup}\label{setup}

\begin{table*}[tb]
\centering
\begin{adjustbox}{max width=\textwidth, keepaspectratio}
\begin{tabular}{ll ccccc ccccc}
\toprule
& & \multicolumn{5}{c}{\textbf{California}} & \multicolumn{5}{c}{\textbf{New York}} \\
\cmidrule(lr){3-7} \cmidrule(lr){8-12}
\textbf{Category} & \textbf{Method} & MAE$\downarrow$ & RMSE$\downarrow$ & MPIW$\downarrow$ & IS$\downarrow$ & COV & MAE$\downarrow$ & RMSE$\downarrow$ & MPIW$\downarrow$ & IS$\downarrow$ & COV \\
\midrule
\multirow{5}{*}{GNN} 
 & DCRNN & 30.537 & 98.941 & 76.215 & 229.083 & \false & 10.957 & 35.371 & 55.028 & 65.715 & \true \\
 & STGCN & 32.708 & 95.427 & 120.883 & 203.678 & \false & 8.967 & 34.218 & 36.705 & 55.320 & \false \\
 & AGCRN & 17.494 & 56.480 & 46.071 & 130.347 & \false & 6.453 & 21.913 & 17.106 & 47.504 & \false \\
 & DGCRN & 28.408 & 84.729 & 96.559 & 208.084 & \false & 17.526 & 57.587 & 19.001 & 273.164 & \false \\
 & UQGNN & 10.235 & 36.520 & 35.680 & 77.672 & \true & 4.585 & 15.820 & 15.230 & 30.426 & \true \\
 \midrule
\multirow{2}{*}{Attention} 
 & DSTAGNN & 20.675 & 55.261 & 63.458 & 131.816 & \false & 8.234 & 29.719 & 32.791 & 49.481 & \false \\
 & ASTGCN & 47.249 & 178.047 & 50.549 & 749.407 & \false & 16.679 & 60.544 & 15.101 & 267.103 & \false \\
\midrule
\multirow{2}{*}{Transformer} 
 & GluonTS & 16.945 & 44.336 & 87.273 & 107.044 & \false & 4.900 & 16.220 & 26.784 & 33.798 & \true \\
 & PatchTST & 12.381 & 36.301 & 59.704 & 80.008 & \false & 5.555 & 20.426 & 29.517 & 37.260 & \true \\
\midrule
\multirow{2}{*}{LLM} 
 & ST-LLM & 14.391 & 47.347 & 37.231 & 111.233 & \false & 5.565 & 20.800 & 16.181 & 38.178 & \false \\
 & UrbanGPT & 9.734 & 38.925 & 47.906 & 83.962 & \true & 4.870 & 17.685 & 20.866 & 31.100 & \true \\
\midrule
\multirow{2}{*}{Mamba} 
 & Mamba & \underline{9.467} & 35.825 & \underline{33.306} & \underline{77.232} & \true & \underline{4.176} & \textbf{14.152} & \underline{14.174} & 30.608 & \false \\
 & U-Mamba & 10.052 & \underline{34.180} & 37.561 & 78.868 & \false & 4.292 & 14.993 & 16.798 & \underline{29.958} & \false \\
\midrule
\rowcolor{gray!20}
Ours & \textbf{\m} & \textbf{9.150} & \textbf{33.401} & \textbf{30.970} & \textbf{74.953} & \true & \textbf{3.926} & \underline{14.395} & \textbf{14.110} & \textbf{29.361} & \true \\
\midrule\midrule

& & \multicolumn{5}{c}{\textbf{Texas}} & \multicolumn{5}{c}{\textbf{Florida}} \\
\cmidrule(lr){3-7} \cmidrule(lr){8-12}
\textbf{Category} & \textbf{Method} & MAE$\downarrow$ & RMSE$\downarrow$ & MPIW$\downarrow$ & IS$\downarrow$ & COV & MAE$\downarrow$ & RMSE$\downarrow$ & MPIW$\downarrow$ & IS$\downarrow$ & COV \\
\midrule
\multirow{5}{*}{GNN} 
 & DCRNN & 25.612 & 178.462 & 34.506 & 344.785 & \false & 185.410 & 418.957 & 403.593 & 856.315 & \false \\
 & STGCN & 22.176 & 164.037 & 75.763 & 165.212 & \false & 168.203 & 385.663 & 489.048 & 786.315 & \false \\
 & AGCRN & 8.632 & 54.573 & 21.883 & 93.392 & \false & 114.045 & 277.833 & 468.500 & 643.579 & \false \\
 & DGCRN & 32.224 & 201.640 & 55.081 & 547.595 & \true & 175.580 & 405.767 & 436.702 & 927.128 & \false \\
 & UQGNN & 7.125 & 49.350 & \textbf{23.580} & 60.850 & \true & 72.350 & 175.620 & 259.559 & 465.230 & \true \\
 \midrule
\multirow{2}{*}{Attention} 
 & DSTAGNN & 13.805 & 91.072 & 49.969 & 95.638 & \false & 150.137 & 355.974 & 447.363 & 654.617 & \false \\
 & ASTGCN & 31.285 & 197.239 & 36.954 & 580.777 & \false & 170.262 & 418.748 & 357.474 & 871.215 & \false \\
\midrule
\multirow{2}{*}{Transformer} 
 & GluonTS & 7.502 & 50.466 & 37.472 & 61.071 & \true & 120.852 & 282.507 & 539.905 & 700.880 & \false \\
 & PatchTST & 28.904 & 191.871 & 25.174 & 450.373 & \false & 167.780 & 370.929 & \textbf{253.671} & 845.201 & \false \\
\midrule
\multirow{2}{*}{LLM} 
 & ST-LLM & 8.921 & 60.350 & 17.686 & 87.761 & \true & 99.501 & 224.992 & 340.592 & 616.386 & \false \\
 & UrbanGPT & 11.024 & 67.689 & 48.997 & 66.529 & \true & \underline{58.459} & \underline{158.116} & 307.648 & \underline{436.149} & \true \\
\midrule
\multirow{2}{*}{Mamba} 
 & Mamba & \underline{6.676} & \underline{47.318} & 24.595 & 61.466 & \false & 85.793 & 209.168 & 386.492 & 567.448 & \false \\
 & U-Mamba & 7.283 & 50.011 & 23.887 & \underline{60.232} & \false & 85.256 & 221.271 & 315.136 & 624.394 & \false \\
\midrule
\rowcolor{gray!20}
Ours & \textbf{\m} & \textbf{6.275} & \textbf{46.566} & \underline{24.015} & \textbf{58.934} & \true & \textbf{54.951} & \textbf{156.535} & \underline{255.230} & \textbf{420.884} & \true \\
\bottomrule
\end{tabular}
\end{adjustbox}
\caption{Comparison with state-of-the-art baselines on four datasets with a horizon of 3 days. $\downarrow$ indicates lower is better. The best results are in \textbf{bold} and the second-best are \underline{underlined}. \true~indicates target coverage ($\geq$90\%) achieved, while \false~indicates failure.}
\label{tab:result}
\end{table*}

\textbf{Datasets.}
We utilize healthcare facility visit data from four major U.S. states (California, New York, Texas, and Florida), provided by Dewey. The dataset captures daily visit counts to four categories (i.e., Ambulatory Health Care Services, Hospitals, Nursing and Residential Care Facilities, and Social Assistance) of healthcare facilities at the county level from January 2019 to April 2022, spanning 40 months.

\noindent \textbf{Baselines.} 
We compare \m with 5 categories of 13 state-of-the-art baselines: 
(1) GNN-based: DCRNN~\cite{dcrnn}, STGCN~\cite{stgcn}, AGCRN~\cite{bai2020adaptive}, DGCRN~\cite{li2023dynamic}, UQGNN~\cite{yu2025uqgnn}; 
(2) Attention-based: DSTAGNN~\cite{lan2022dstagnn}, ASTGCN~\cite{zhu2021ast}; 
(3) Transformer-based: GluonTS~\cite{alexandrov2020gluonts}, PatchTST~\cite{nie2022time}; 
(4) LLM-based: ST-LLM~\cite{liu2024st}, UrbanGPT~\cite{li2024urbangpt};
(5) Mamba-based: Mamba~\cite{gu2024mamba}, U-Mamba~\cite{ma2024u}.

\noindent \textbf{Metrics.} 
We utilize two widely used metrics, including Mean Absolute Error (MAE) and Root Mean Squared Error (RMSE), to evaluate the performance of deterministic prediction, and three other commonly used metrics, including Mean Prediction Interval Width (MPIW), Interval Score (IS), and Coverage (COV), to evaluate the performance of uncertainty quantification. 
It is worth noting that our model outputs prediction intervals, and the deterministic metrics are computed based on the mean of the upper and lower bounds.

\subsection{Overall Performance Comparison (RQ 1)}
\begin{figure}[tb]
\centering
\subfigure[UrbanGPT vs. HealthMamba]{
\label{pres2}
\includegraphics[width=0.48\linewidth]{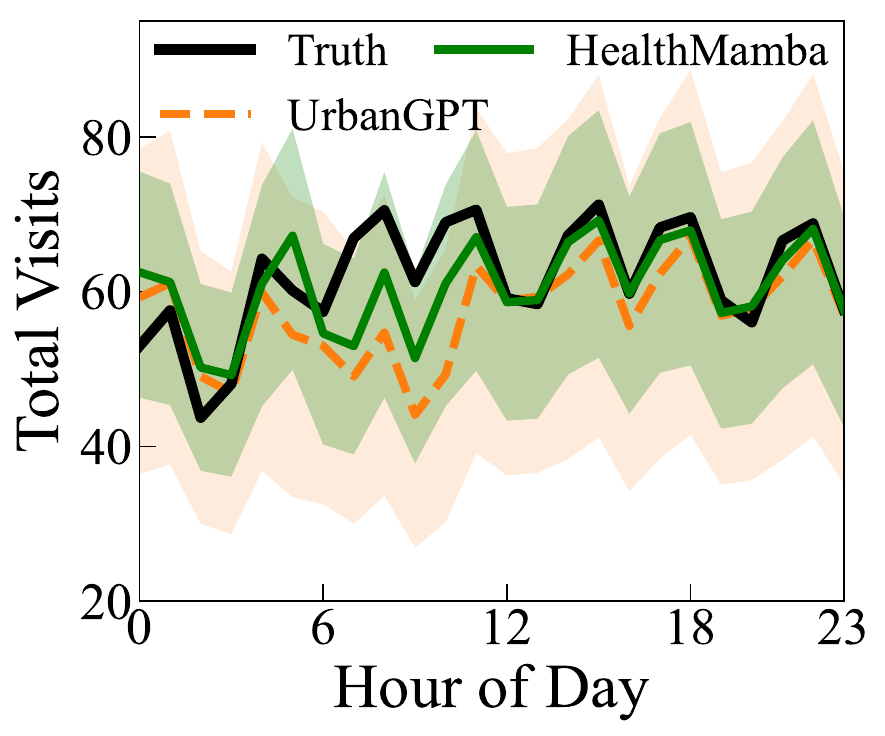}}
\subfigure[Mamba vs. HealthMamba]{
\label{pres3}
\includegraphics[width=0.48\linewidth]{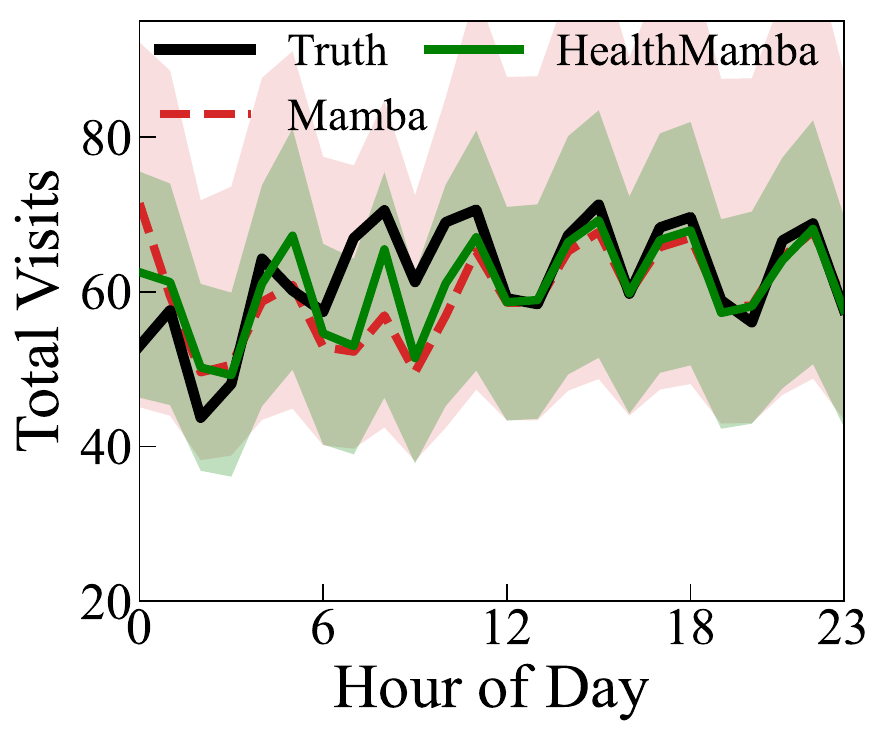}}
\caption{Prediction results on the California dataset aggregated in 24 hours, with best baselines selected for visualization. }
\label{probres}
\end{figure}

An overall comparison of our \m and other baselines is presented in Table \ref{tab:result}. We found that our \m 
reduces MAE by approximately 6.0\% compared to the best baseline Mamba based on the average of all four datasets.
In addition, \m also demonstrates superior performance on prediction reliability, with a 3.5\% improvement in IS and reaching the target coverage. 
As shown in Figure~\ref{probres}, the green shadow of \m is more compact while still covering most observations, indicating more reliable and precise estimates compared to baselines.
Our experiments show that \m consistently achieves the best performance for both short-term (next-day) and medium-term (7-day-ahead) prediction.

\subsection{Type-specific Visit Prediction (RQ 2)}
\begin{table}[tb]\small
\centering
\setlength{\tabcolsep}{2.0pt}
\begin{tabular}{@{}llccccc@{}}
\toprule
\textbf{Category} & \textbf{Model} & MAE$\downarrow$ & RMSE$\downarrow$ & MPIW$\downarrow$ & IS$\downarrow$ & COV \\
\midrule
\multirow{3}{*}{Hospitals} & Mamba & 12.384 & 42.156 & 38.720 & 95.832 & \true \\
& UrbanGPT & 12.058 & 45.623 & 52.418 & 101.254 & \true \\
& \m & \textbf{11.628} & \textbf{39.524} & \textbf{35.105} & \textbf{91.250} & \true \\
\midrule
\multirow{3}{*}{Ambulatory} & Mamba & 8.152 & 31.478 & 29.640 & 68.425 & \true \\
& UrbanGPT & 8.536 & 34.127 & 42.815 & 73.680 & \true \\
& \m & \textbf{7.683} & \textbf{29.352} & \textbf{27.218} & \textbf{65.127} & \true \\
\midrule
\multirow{3}{*}{Nursing} & Mamba & 7.945 & 28.634 & 31.852 & 62.148 & \false \\
& UrbanGPT & 8.213 & 32.576 & 45.037 & 70.425 & \true \\
& \m & \textbf{7.486} & \textbf{26.815} & \textbf{29.476} & \textbf{59.268} & \true \\
\midrule
\multirow{3}{*}{Social} & Mamba & 9.387 & 40.032 & 33.012 & 82.523 & \false \\
& UrbanGPT & 10.127 & 43.374 & 51.352 & 89.489 & \true \\
& \m & \textbf{8.803} & \textbf{37.913} & \textbf{31.081} & \textbf{78.167} & \true \\
\bottomrule
\end{tabular}
\caption{Fine-grained prediction on four healthcare facility types.}
\label{tab:granularity}
\end{table}

We further show the prediction results on different healthcare facility categories in Table~\ref{tab:granularity}. We found that \m consistently outperforms Mamba and UrbanGPT across all categories. Notably, Mamba fails to achieve target coverage in the Nursing and Social categories, while UrbanGPT exhibits substantially wider prediction intervals despite achieving coverage. In contrast, \m maintains valid and compact confidence intervals across all categories, demonstrating robustness to facility-type heterogeneity.

\subsection{Ablation Study (RQ 3)}
To evaluate the contribution of each key component in \m, we conduct an ablation study by comparing the full model against six variants: 
(1) \textbf{w/o STCE}, which removes the contextual information; 
(2) \textbf{w/o G-Mamba}, which replaces the GraphMamba with conventional Mamba; 
(3) \textbf{w/o Node-based} UQ module, (4) \textbf{w/o Distribution-based} UQ module, and (5) \textbf{w/o Parameter-based} UQ module; and 
(6) \textbf{w/o UQ}, which uses only quantile regression.

\begin{table}[tb]\small

\centering
\setlength{\tabcolsep}{3pt}
\begin{tabular}{@{}lccccc@{}}
\toprule
\textbf{Variants} & MAE$\downarrow$ & RMSE$\downarrow$& MPIW$\downarrow$ & IS$\downarrow$ & COV \\ 
\midrule
w/o STCE & 13.610 & 48.535 & 39.807 & 108.110 & \false \\ 
w/o G-Mamba & 15.782 & 49.092 & 42.625 & 117.382 & \false \\
\midrule
w/o Node-based & 9.682 & 34.925 & 32.540 & 80.115 & \false\\ 
w/o Distribution-based & 10.154 & 36.218 & 34.112 & 85.330 & \false \\ 
w/o Parameter-based & 10.573 & 37.665 & 35.890 & 90.224 & \false \\ 
w/o UQ (Total) & 11.240 & 39.150 & 37.520 & - & - \\ 
\midrule
\textbf{\m} & \textbf{9.150} & \textbf{33.401} & \textbf{30.970} & \textbf{74.953}  & \true \\ 
\bottomrule
\end{tabular}
\caption{Ablation study on the California dataset.}
\label{tab:ablation}
\end{table}


Table \ref{tab:ablation} validates our design choices. Architecturally, removing GraphMamba or STCE significantly degrades MAE (up to 15.782), highlighting the crucial role of spatiotemporal and contextual modeling. Furthermore, omitting any individual UQ module (Node-, Parameter-, or Distribution-based) increases MAE and fails the coverage constraint (COV = \false). Finally, completely removing UQ fails to provide valid prediction intervals entirely (missing IS and COV), demonstrating the necessity of our synergistic UQ framework.

\subsection{Prediction under Abnormal Situations (RQ 4)}

\begin{figure}[tb]
\centering
\subfigure[California: COVID-19.]{
\includegraphics[width=0.48\linewidth]{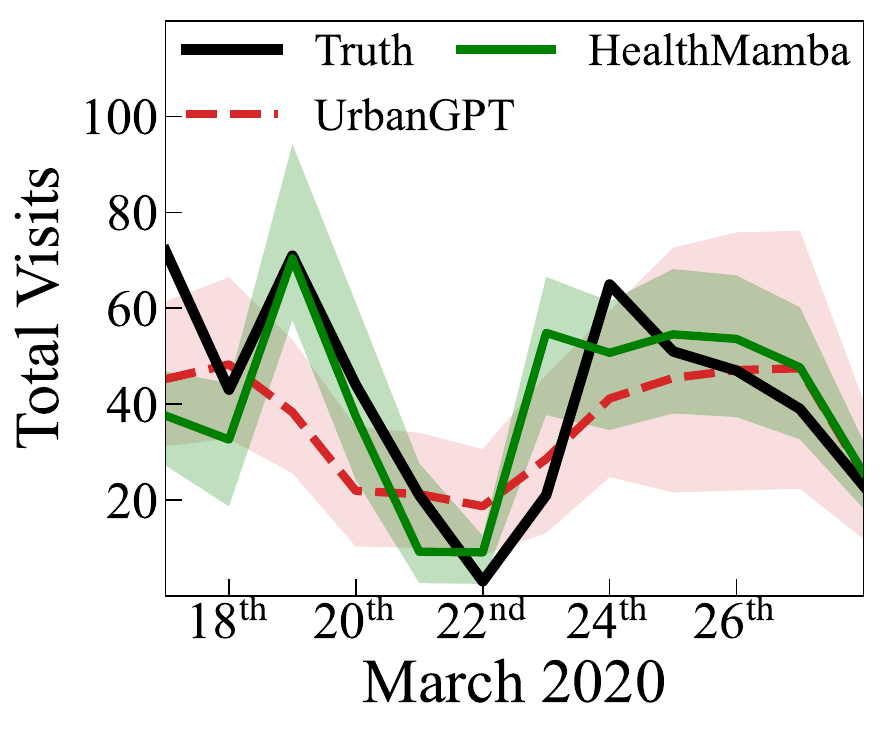}}
\subfigure[Texas: Hurricane Hanna.]{
\includegraphics[width=0.48\linewidth]{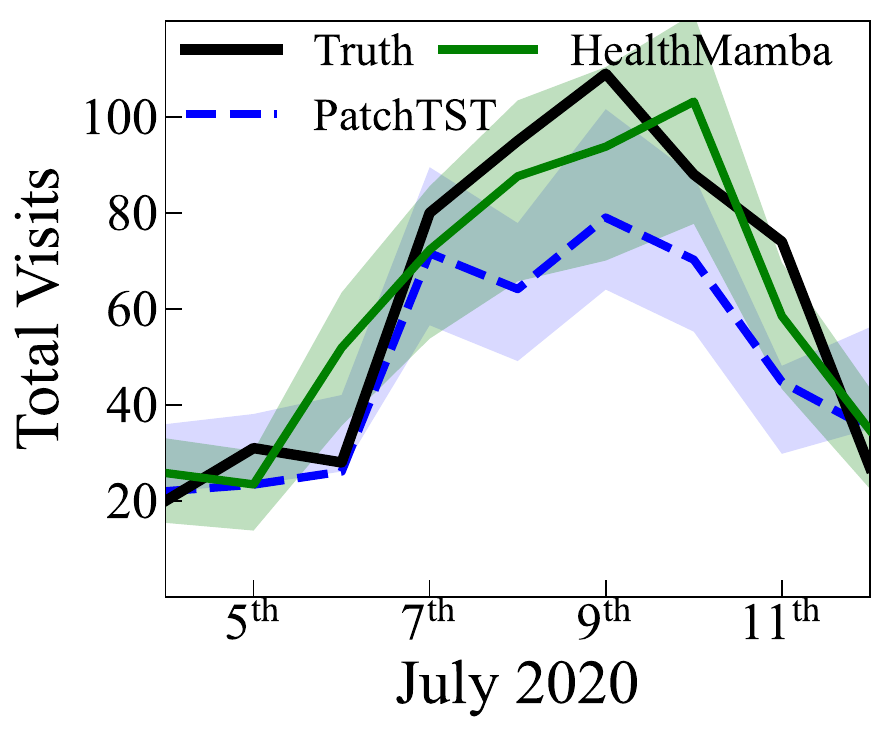}}
\caption{Prediction results under extreme events.}
\label{anomaly}
\end{figure}

A key challenge in healthcare visit prediction is handling abnormal scenarios (e.g., emergencies, extreme weather) that cause sudden pattern deviations. We evaluate \m on two representative events: the COVID-19 lockdown in LA County (March 2020) causing a sudden drop in visits, and Hurricane Hanna in Texas (July 2020) triggering a surge. As shown in Figure~\ref{anomaly}, under these situations, \m successfully maintained accurate predictions, and ground truth values are also within the prediction intervals in most cases, whereas the best baseline struggled to adapt to these abrupt changes. These results highlight \m's superior adaptability to both demand surges and drops under abnormal scenarios.


\section{Related Work}\label{literature review}

\subsection{Healthcare Facility Visit Prediction}
Healthcare facility visit prediction has attracted much interest from both academia and the government, as it is important for health resource allocation and public emergency responses. 
Traditional methods rely on statistical models such as ARIMA~\cite{orhan2025predicting} and regression-based methods~\cite{avinash2025time,shen2025learning} for daily or weekly patient count prediction, but overlook temporal dynamics and spatial dependencies.
Recent works have explored different approaches to address this problem,
including GNN-based methods~\cite{stgcn,10446624,shen2026cited},
attention-based methods~\cite{lan2022dstagnn},
Transformer-based methods~\cite{nie2022time}, 
and LLM-based methods~\cite{li2024urbangpt,li2026llmclinicalgraphstructure,10.1145/3748636.3763223}.
However, most existing works focus on aggregated prediction without considering different types of healthcare facilities, such as hospitals and nursing facilities, which is more important for practical decision support. In addition, existing solutions usually consider this problem a time-series prediction problem, without capturing spatial dependencies, which can potentially be used to improve prediction performance.

\subsection{Uncertainty-aware Spatiotemporal Prediction}
Uncertainty quantification has attracted much interest from the spatiotemporal prediction community because it lays the foundation for reliable and safe decision-making~\cite{jiang2025uncertainty,yu2026trustenergy,cheng2025bts}.
DeepSTUQ~\cite{qian2023uncertainty} employs dual sub-networks to separately quantify aleatoric and epistemic uncertainty at each node. 
DiffSTG~\cite{wen2023diffstg} generalizes denoising diffusion probabilistic models to spatiotemporal graphs to capture intrinsic uncertainties. STZINB-GNN~\cite{stzinb-gnn} models sparse spatiotemporal data through zero-inflated negative binomial distributions. CF-GNN~\cite{huang2024uncertainty} extends conformal prediction to graph-based models for guaranteed coverage, providing distribution-free prediction intervals.
UQGNN~\cite{yu2025uqgnn} integrates graph neural networks with node-level uncertainty estimation for spatiotemporal prediction.
Different from existing works~\cite{yang2023re,yang2024regulating}, we design a comprehensive uncertainty quantification module that integrates distribution-based, parameter-based, and node-based uncertainty quantification mechanisms, combined with post-hoc quantile calibration to produce well-calibrated prediction intervals for uncertainty-aware spatiotemporal prediction.
\section{Conclusion}\label{conclusion}
In this paper, we propose \m, an uncertainty-aware spatiotemporal framework for accurate and reliable healthcare facility visit prediction. There are three key components in \m: (i) a Unified Spatiotemporal Context Encoder that integrates both heterogeneous static and dynamic contextual information, (ii) a novel graph state space model called GraphMamba for spatiotemporal modeling, and (iii) a comprehensive UQ module integrating three different types of UQ mechanisms and a post-hoc quantile calibration for higher prediction reliability.
Extensive experiments on four real-world datasets from California, New York, Texas, and Florida demonstrate that \m outperforms 13 state-of-the-art baselines, improving prediction accuracy by 6.0\% and uncertainty quantification by 3.5\%. Our uncertainty-aware design also enables robust predictions under public emergencies and extreme weather events, supporting trustworthy decision-making under abnormal scenarios.

\clearpage
\newpage

\section*{Acknowledgments}
We sincerely thank all anonymous reviewers for their insightful comments and valuable suggestions. This work is partially supported by the National Science Foundation under Grant 2411152 and the FSU Institute for Successful Longevity (ISL) Planning Grant.

\section*{Ethics Statement}
All authors comply with the IJCAI Code of Ethics and take full responsibility for this research. We use only de-identified data without human subjects or sensitive information, posing no foreseeable harm. There are no conflicts of interest or external sponsorships.

\bibliographystyle{named}
\bibliography{reference}

\newpage
\appendix
\section*{Appendix}

\section{Pseudocodes}\label{app_framework}

\begin{algorithm}[H]
\caption{STCE: Produce embedding $\mathbf{R}$ from inputs}
\begin{algorithmic}[1]
\STATE \textbf{Input:} $\mathbf{V}\in\mathbb{R}^{N\times T_{\mathrm{in}}\times C}$, $\mathbf{D}\in\mathbb{R}^{N\times d_{\mathrm{dem}}}$, $\mathbf{E}\in\mathbb{R}^{N\times T_{\mathrm{in}}\times d_{\mathrm{ext}}}$, $\mathbf{A}\in\mathbb{R}^{N\times N}$
\FOR{each node $i$ and time $t$}
  \STATE $\mathbf{H}^v_{i,t}, \mathbf{H}^d_i, \mathbf{H}^e_{i,t} \leftarrow$ FeatureEmbeddingLayers$(\mathbf{v}_{i,t}, \mathbf{D}_i, \mathbf{E}_{i,t})$
  \STATE $\mathbf{X}_{i,t} \leftarrow \mathbf{H}^v_{i,t} + \mathbf{H}^d_i + \mathbf{H}^e_{i,t}$
\ENDFOR
\FOR{each time $t$}
  \STATE $\widetilde{\mathbf{X}}_{:,t} \leftarrow \mathrm{GConv}(\mathbf{X}_{:,t}; \mathbf{A})$
\ENDFOR
\STATE $\widehat{\mathbf{X}} \leftarrow \mathrm{TemporalMixer}(\widetilde{\mathbf{X}})$
\STATE $\mathbf{R} \leftarrow \mathrm{LayerNorm}(\mathrm{Proj}([\widehat{\mathbf{X}}; \mathbf{H}^d]))$
\RETURN $\mathbf{R}\in\mathbb{R}^{N\times T_{\mathrm{in}}\times d_{\mathrm{model}}}$
\end{algorithmic}
\end{algorithm}

\begin{algorithm}[H]
\caption{GraphMamba Backbone Forward Pass}
\begin{algorithmic}[1]\label{algo2}
\STATE \textbf{Input:} $\mathbf{X}^{(0)} = \mathbf{R}$, prior adjacency $\mathbf{A}_0$

\STATE \textbf{/* Encoder */}
\FOR{$s = 0$ \textbf{to} $S-1$}
    \STATE $\mathbf{Y}^{(s)} \leftarrow \text{G-Mamba}^{(s)}\!\big(\mathbf{X}^{(s)}; \mathbf{A}^{\ast}\big)$
    \STATE $\mathbf{X}^{(s+1)} \leftarrow \mathrm{DownConv}\!\big(\mathbf{Y}^{(s)}\big)$
\ENDFOR

\STATE \textbf{/* Bottleneck */}
\STATE $\mathbf{Y}^{(S)} \leftarrow \text{G-Mamba}^{(S)}\!\big(\mathbf{X}^{(S)}; \mathbf{A}^{\ast}\big)$
\STATE $\mathbf{Z}^{(S)} \leftarrow \mathbf{Y}^{(S)}$

\STATE \textbf{/* Decoder */}
\FOR{$s = S-1$ \textbf{down to} $0$}
    \STATE $\widehat{\mathbf{Y}}^{(s)} \leftarrow \mathrm{UpConv}\!\big(\mathbf{Z}^{(s+1)}\big)$
    \STATE $\mathbf{Z}^{(s)} \leftarrow \text{G-Mamba}^{(s)}\!\Big(
        \mathrm{Fuse}\big(\mathbf{Y}^{(s)},\widehat{\mathbf{Y}}^{(s)}\big); \mathbf{A}^{\ast}
    \Big)$
\ENDFOR

\STATE \textbf{Output:} Multi-scale representations $\{\mathbf{Z}^{(s)}\}_{s=0}^{S}$
\end{algorithmic}
\end{algorithm}

\section{Experiment Setup}\label{app_experiment}

\subsection{Details and Parameters of Baselines}\label{app_baseline}

\subsubsection{GNN-based Methods}\label{app_gnn_methods}
\begin{itemize}[noitemsep, left=0pt]
\item \textbf{DCRNN}~\cite{dcrnn}: Diffusion Convolutional Recurrent Neural Network integrates diffusion convolution with a sequence-to-sequence architecture to learn representations of spatial dependencies and temporal relations.
We set the diffusion step to 2 and use a 2-layer seq2seq GRU with hidden size 64; scheduled sampling is enabled and the dropout rate is set to 0.3.
\item \textbf{STGCN}~\cite{stgcn}: Spatial-Temporal Graph Convolution Network combines spectral graph convolution with 1D convolution to capture spatial and temporal correlations.
The spatial and temporal kernel size is set to 3, the block number is set to 2, and the dropout rate is set to 0.5.
\item \textbf{AGCRN}~\cite{bai2020adaptive}: An Adaptive Graph Convolutional Recurrent Network that uses a Node Adaptive Parameter Learning module to capture node-specific patterns and a Data Adaptive Graph Generation module to infer inter-dependencies among different traffic series.
The order of Chebyshev polynomials is set to 2, the number of blocks is 2, and the RNN unit dimension is set to 64.
\item \textbf{DGCRN}~\cite{li2023dynamic}: A Dynamic Graph Convolutional Recurrent Network that employs hyper-networks to dynamically generate graph convolutional filters at each time step, capturing evolving spatial dependencies by combining a learned dynamic graph with a pre-defined static road network.
We use 2 DGCRN blocks with hidden size 64, set the Chebyshev polynomial order to 2, and apply dropout with rate 0.3.
\item \textbf{UQGNN}~\cite{yu2025uqgnn}: An Uncertainty Quantification Graph Neural Network for multivariate spatiotemporal prediction that introduces an interaction-aware spatiotemporal embedding module along with a multivariate probabilistic prediction module to estimate both expected mean values and associated uncertainties.
We use 2 diffusion steps with hidden size 64, set the number of TCN layers to 2 with kernel size 3, and apply dropout with rate 0.3.

\end{itemize}

\subsubsection{Attention-based Methods}
\begin{itemize}[noitemsep, left=0pt]
\item \textbf{DSTAGNN}~\cite{lan2022dstagnn}: A Dynamic Spatial-Temporal Aware Graph Neural Network that constructs a data-driven, time-varying graph to replace the static topology, using multi-head attention with multi-scale gated convolutions to capture complex spatial and temporal dependencies in traffic flow.
We set the number of attention heads to 4, use 2 stacked DSTAGNN layers with hidden size 64, and set the dropout rate to 0.3.
\item \textbf{ASTGCN}~\cite{guo2021learning}: An Attention-based Spatio-Temporal Graph Convolutional Network that employs spatial-temporal attention within graph convolutions, using separate modules for recent, daily-periodic, and weekly-periodic patterns to effectively model dynamic correlations in traffic data.
The order of Chebyshev polynomials is set to 2, the number of blocks is also 2, and the time stride is set to 1.
\end{itemize}
\subsubsection{Transformer-based Methods}
\begin{itemize}[noitemsep, left=0pt]
\item \textbf{GluonTS}~\cite{alexandrov2020gluonts}: A toolkit for probabilistic time series forecasting that provides deep learning-based models, reference implementations of state-of-the-art methods, and utilities for efficient experimentation and evaluation.
We use its probabilistic forecasting backbone with 2 recurrent layers (hidden size 64) and set the context window length to 168; the dropout rate is set to 0.1.
\item \textbf{PatchTST}~\cite{nie2022time}: A Transformer-based model for multivariate time series forecasting that segments each series into subseries-level patches and processes each channel independently, enabling long-range attention and state-of-the-art long-term forecasting performance.
We set the patch length to 6 with stride 2, use 3 Transformer encoder layers with 4 attention heads, and set the model dimension to 64 with dropout 0.1.
\end{itemize}

\subsubsection{LLM-based Methods}
\begin{itemize}[noitemsep, left=0pt]
\item \textbf{ST-LLM}~\cite{liu2024st}: A Spatial-Temporal Large Language Model for traffic prediction that treats each location's time series as a sequence of tokens with learned spatial-temporal embeddings, using a partially frozen attention mechanism to capture global spatio-temporal patterns.
We adopt a partially frozen LLM and train lightweight adapters; the spatiotemporal embedding dimension is set to 64, and the context length is set to 512 tokens.
\item \textbf{UrbanGPT}~\cite{li2024urbangpt}: A spatio-temporal LLM framework that integrates a dependency encoder with instruction-tuning, enabling the model to learn complex inter-dependencies across time and space and to generalize to diverse urban prediction tasks, especially under data scarcity.
We follow the instruction-tuning setup with a dependency encoder of hidden size 64 and 2 layers; the LLM backbone is kept frozen, and the adapter dropout rate is set to 0.1.
\end{itemize}

\subsubsection{Mamba-based Methods}
\begin{itemize}[noitemsep, left=0pt]
\item \textbf{Mamba}~\cite{gu2024mamba}: A selective state space model (SSM) architecture that enables efficient long-sequence modeling with linear-time complexity, serving as a strong alternative to attention for capturing long-range temporal dependencies.
We use 4 stacked Mamba blocks with model dimension 64 and set the state expansion factor to 2; dropout is set to 0.1.
\item \textbf{U-Mamba}~\cite{ma2024u}: A U-shaped Mamba architecture that combines hierarchical encoder-decoder processing with skip connections, allowing multi-scale feature extraction while retaining long-range dependency modeling through Mamba blocks.
We use a 2-level encoder-decoder with 2 Mamba blocks per stage, set the base width to 64, and apply dropout with rate 0.1.
\end{itemize}

\subsection{Metrics}\label{app_metrics}
\subsubsection{Deterministic Metrics}
We utilize two commonly used deterministic metrics, including Mean Absolute Error (MAE) and Root Mean Squared Error (RMSE), to evaluate the performance of deterministic predictions: 
\begin{equation}
    \begin{aligned}
&\text{MAE}=\frac{1}{N} \sum_{i=1}^{N}|y_i-\hat{y}_i|,
    \end{aligned}
\end{equation}
\begin{equation}
    \begin{aligned}
    &\text{RMSE} = \sqrt{\frac{1}{N}\sum_{i=1}^{N}(y_i - \hat{y}_i)^2},
    \end{aligned}
\end{equation}
where $y_i$ is the ground truth label, $\hat{y}_i$ denotes the prediction result, and $N$ is the number of predictions.

\subsubsection{Probabilistic Metrics}
We also utilize three other metrics, including Mean Prediction Interval Width (MPIW), Interval Score (IS), and Coverage, to evaluate the performance of uncertainty quantification.

MPIW quantifies the average width of prediction intervals a model generates to capture uncertainty in its predictions. Formally, MPIW is defined as:
\begin{equation}
    \begin{aligned}
    \text{MPIW}=\frac{1}{N} \sum_{i=1}^{N}(u_{\alpha,i} - l_{\alpha,i}),
    \end{aligned}    
\end{equation}
where $l_{\alpha, i}$ and $u_{\alpha, i}$ are the lower and upper quantile for the $i$-th prediction under the miscoverage $\alpha$. Smaller values indicate narrower intervals. MPIW is often used alongside coverage metrics to ensure the prediction intervals balance informativeness and reliability.

It is also important to evaluate the quality of the prediction interval, in addition to calculating the length. 
Interval Score (IS) evaluates both the width and coverage quality of prediction intervals by penalizing intervals that are too wide or fail to contain the true value. Formally, for miscoverage level $\alpha$, IS is defined as:
\begin{equation}
\begin{aligned}
\text{IS} = \frac{1}{N} \sum_{i=1}^{N} \Big[ (u_{\alpha,i} - l_{\alpha,i})
&+ \frac{2}{\alpha}\,(l_{\alpha,i} - y_i)\,\mathbb{I}(y_i < l_{\alpha,i}) \\
&+ \frac{2}{\alpha}\,(y_i - u_{\alpha,i})\,\mathbb{I}(y_i > u_{\alpha,i}) \Big],
\end{aligned}
\end{equation}
where $l_{\alpha,i}$ and $u_{\alpha,i}$ denote the lower and upper bounds of the prediction interval. A smaller values indicate narrower intervals, reflecting higher model confidence.

In addition, we also need to know how the prediction interval covers the ground truth. 
Given $N$ samples, the prediction interval is denoted as $[l_i,u_i], i=1,2,...N$ and the ground truth is expressed as $y_i, i=1,2,...N$.
The coverage is calculated according to:
\begin{equation}
\text{coverage}=\frac{100\%}{N}\sum_{i=1}^{N}\mathbb{I}(l_i<y_i<u_i).
\end{equation}

\begin{table}[tb]\small

\centering
\setlength{\tabcolsep}{7pt}
\begin{adjustbox}{max width=0.48\textwidth, keepaspectratio}
\begin{tabular}{@{}ll ccc@{}}
\toprule
\textbf{Category} & \textbf{Method} & Train$\downarrow$ & Mem$\downarrow$ & Params$\downarrow$ \\
\midrule
\multirow{5}{*}{GNN} 
 & DCRNN    & 83 & 1,494 & 25  \\
 & STGCN    & 31 & 1,621 & 55  \\
 & AGCRN    & 135 & 1,567 & 757  \\
 & DGCRN    & 128 & 1,539 & 251  \\
 & UQGNN    & 95 & 1,580 & 186  \\
\midrule
\multirow{2}{*}{Attention} 
 & DSTAGNN  & 720 & 1,709 & 110  \\
 & ASTGCN   & 72 & 1,679 & 66  \\
\midrule
\multirow{2}{*}{Transformer} 
 & GluonTS  & 70 & 1,660 & 28  \\
 & PatchTST & 35 & 1,470 & 77  \\
\midrule
\multirow{2}{*}{LLM} 
 & ST-LLM   & 1522 & 4,280 & 124,439  \\
 & UrbanGPT & 2063 & 14,336 & 6,738,415  \\
\midrule
\multirow{2}{*}{Mamba} 
 & Mamba    & 42 & 1,700 & 664  \\
 & U-Mamba  & 58 & 1,715 & 804  \\
\midrule
Ours & \m & 65 & 1,520 & 892 \\
\bottomrule
\end{tabular}
\end{adjustbox}
\caption{Computational complexity comparison on the California dataset. Train denotes total training time (seconds). Mem denotes peak GPU memory usage (MB). Params denotes the number of trainable parameters (K).}
\label{tab:complexity}
\end{table}

\subsection{Implementation Details}\label{app_implementation}
All experiments were conducted on a Linux platform equipped with an NVIDIA A100 GPU with 80 GB of memory.
For training, we use the Adam optimizer with a batch size of 128. The initial learning rate is set to \(1 \times 10^{-3}\), with a decay rate of \(5 \times 10^{-4}\), applied every 15 epochs. We employ early stopping with a patience of 50 steps based on the validation loss to prevent overfitting. 
The dataset is divided into training, validation, and testing sets in a ratio of 8:1:1. The input sequence length is set to 7 time steps (7 days), and the model's output horizon is 3. All time series data are normalized using a transformation of natural logarithm, which is represented as:
\begin{equation}
    X'=\ln(X+1),
\end{equation}
where $X$ denotes the original dataset and $X'$ is the normalized dataset. All values are incremented by 1 to avoid undefined logarithms for zero-valued entries.

For our proposed \m, the model dimension $d_{\mathrm{model}}$ is set to 64, and the hidden dimension $d_{\mathrm{hid}}$ is set to 128.
In the STCE module, the demographic embedding dimension $d_{\mathrm{dem}}$ is 32, and the external factor dimension $d_{\mathrm{ext}}$ is 16.
For the GraphMamba backbone, we use $S=2$ encoder-decoder stages with 2 Mamba blocks per stage; the state expansion factor is set to 2, and the convolutional kernel size is 4.
In the Uncertainty-aware Prediction Head, the target miscoverage rate $\alpha$ is set to 0.1 (corresponding to 90\% coverage).
Dropout is applied with a rate of 0.3, and gradient clipping is set to 1.0.

\section{Additional Evaluation}\label{app_evaluation}

\subsection{Detailed Complexity Analysis}\label{app_complexity}
We analyze the computational complexity of \m compared to baselines in terms of training time per epoch, peak GPU memory usage, and the number of parameters. As presented in Table~\ref{tab:complexity}, our approach achieves a favorable balance between model capacity and computational efficiency.

In terms of training time, \m requires 65 seconds per epoch, which is comparable to lightweight models like GluonTS (70s) and ASTGCN (72s), and significantly faster than attention-heavy models like DSTAGNN (720s). Notably, LLM-based methods exhibit substantially higher training costs, with ST-LLM requiring 1,522 seconds and UrbanGPT requiring 2,063 seconds per epoch.

For GPU memory consumption, \m uses only 1,520 MB, which is even lower than basic models like STGCN (1,621 MB) and the vanilla Mamba (1,700 MB). This efficient memory footprint is particularly advantageous for deployment scenarios with limited GPU resources. In contrast, LLM-based approaches demand substantially more memory, with UrbanGPT requiring 14,336 MB.

Regarding parameter count, \m contains 892K trainable parameters, which is moderate compared to GNN-based methods (ranging from 25K for DCRNN to 757K for AGCRN) and similar to Mamba-based baselines (664K--804K). This parameter efficiency, combined with comparable or superior prediction performance, suggests that our architecture effectively utilizes its capacity. The LLM-based methods contain orders of magnitude more parameters (124M for ST-LLM and 6.7B for UrbanGPT), yet do not achieve proportionally better results, highlighting the efficiency of our specialized design for healthcare facility visit prediction.

\begin{figure}[tb]
\centering
\subfigure[Selective regression]{\label{curve1}
\includegraphics[width=0.48\linewidth]{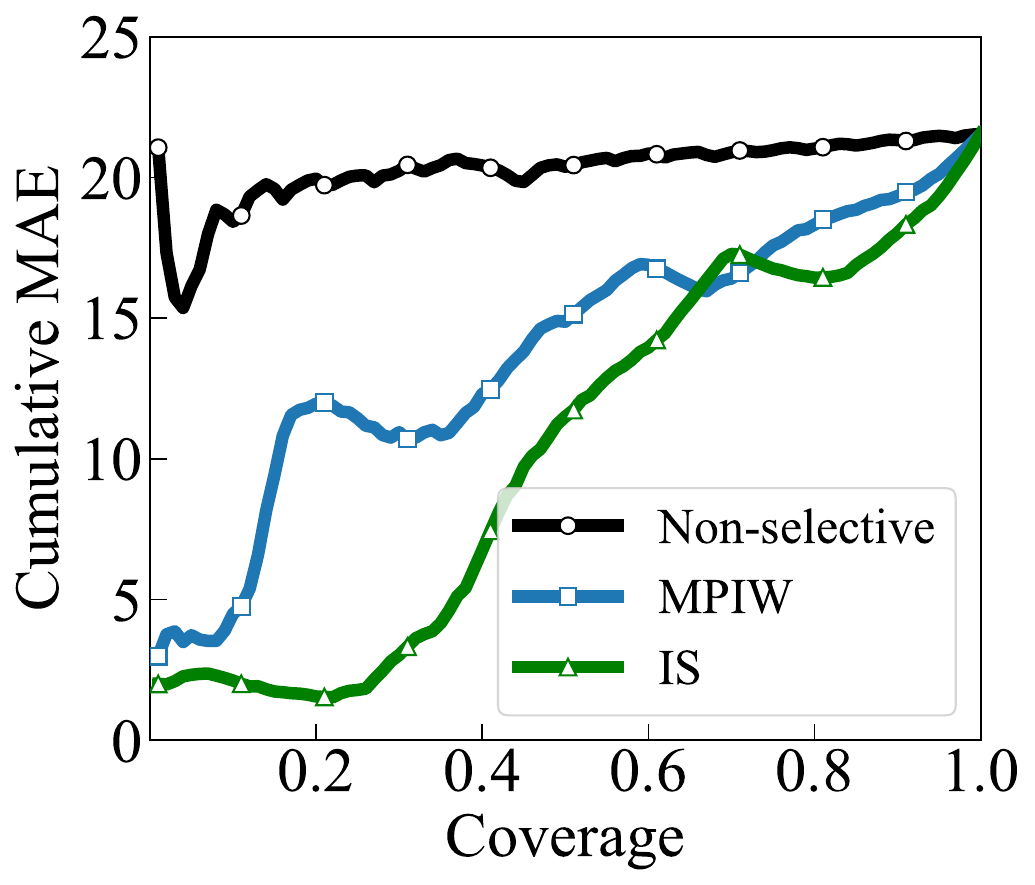}}
\subfigure[Ideal vs. Empirical Coverage]{\label{curve2}
\includegraphics[width=0.48\linewidth]{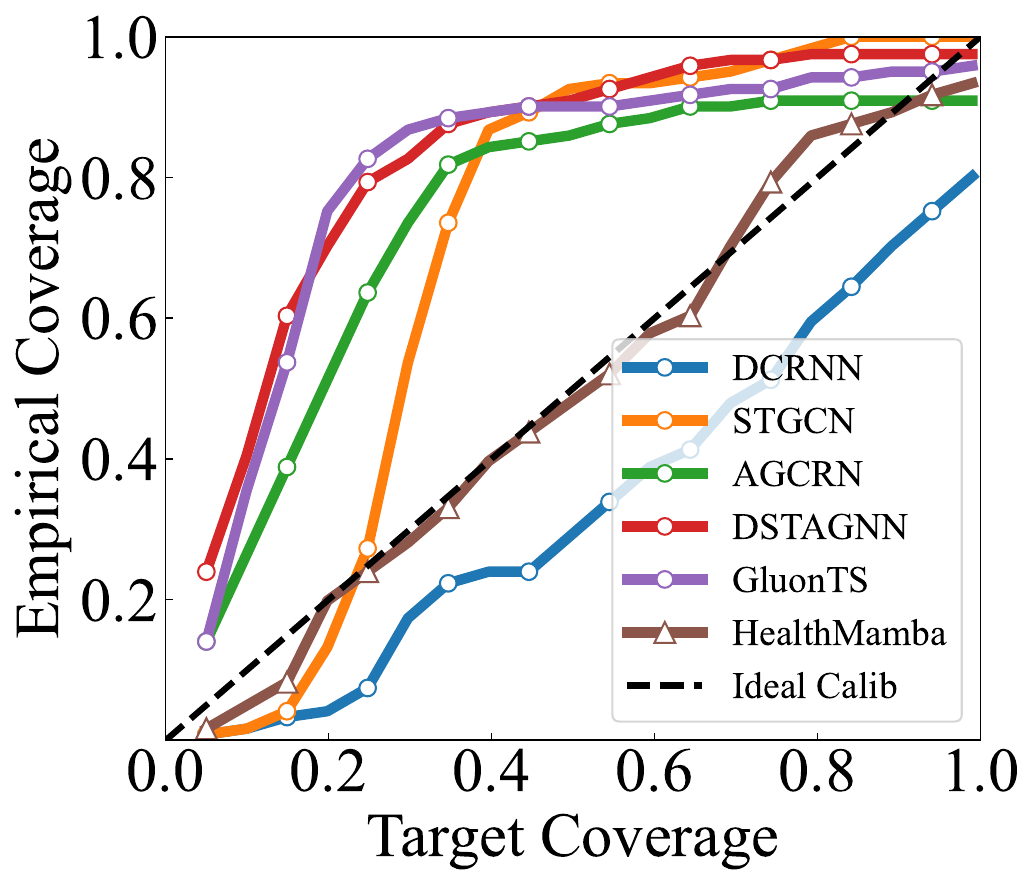}}
\caption{UQ analyses on the California dataset.}
\label{curve_figure}
\end{figure}

\subsection{Effectiveness of UQ}
To further validate the effectiveness of our uncertainty quantification, we employ selective regression~\cite{sokol2024conformalized,shah2022selective}, which enables the model to abstain from predictions when confidence is insufficient.
As illustrated in Figure~\ref{curve1}, the MAE curve remains nearly flat when uncertainty quantification is not used, whereas the error consistently increases with coverage once uncertainty estimates are incorporated. This confirms that the estimated uncertainty scores are meaningful.
Besides, Figure~\ref{curve2} shows \m achieves the most reliable calibration as its curve is closest to the diagonal line.

\subsection{Prediction under Different Horizons}\label{app_horizon}
In practice, decision-makers such as governments and public health agencies require healthcare facility visit predictions at different temporal horizons to support effective policy planning. For example, short-term next-day predictions are critical for real-time resource allocation and emergency response, while medium-term (one-week) predictions can inform staffing and intervention design. The ability to provide accurate and reliable predictions across different forecasting horizons is therefore essential for translating predictive models into actionable public health decisions. Hence, we also evaluate \m across different prediction horizons (1 and 7 days) to assess both short-term and medium-term forecasting capability. As shown in Table~\ref{tab:horizon_1} and Table~\ref{tab:horizon_2}, our model consistently outperforms baselines across all four datasets under both horizons.

For short-term (next-day) predictions, \m achieves the lowest MAE on all datasets, demonstrating superior point prediction accuracy. Notably, our model also maintains the best Interval Score (IS) across most datasets, indicating well-calibrated uncertainty estimates with appropriately narrow prediction intervals.
For medium-term (7-day-ahead) predictions, the performance gap between \m and baselines becomes more pronounced. While all methods experience degradation as the horizon extends, \m exhibits the most graceful degradation. For instance, on the California dataset, our MAE increases from 6.78 to 10.84, compared to Mamba's increase from 7.17 to 11.44 with much larger absolute errors. More importantly, \m is the only method that consistently achieves the target 90\% coverage across all datasets for both horizons, validating the robustness of our uncertainty quantification mechanisms.

\begin{table*}[tb]\small

\centering
\begin{adjustbox}{max width=\textwidth, keepaspectratio}
\begin{tabular}{@{}ll ccccc ccccc@{}}
\toprule
& & \multicolumn{5}{c}{\textbf{California}} & \multicolumn{5}{c}{\textbf{New York}} \\
\cmidrule(lr){3-7} \cmidrule(lr){8-12}
\textbf{Category} & \textbf{Method} & MAE$\downarrow$ & RMSE$\downarrow$ & MPIW$\downarrow$ & IS$\downarrow$ & COV & MAE$\downarrow$ & RMSE$\downarrow$ & MPIW$\downarrow$ & IS$\downarrow$ & COV \\
\midrule
\multirow{5}{*}{GNN} 
 & DCRNN & 23.14 & 74.89 & 58.23 & 172.45 & \false & 8.23 & 26.78 & 41.52 & 49.33 & \false \\
 & STGCN & 24.81 & 72.25 & 91.88 & 154.17 & \false & 6.78 & 25.91 & 27.85 & 41.89 & \false \\
 & AGCRN & 13.24 & 42.76 & 34.89 & 98.68 & \true & 4.88 & 16.59 & 12.94 & 35.90 & \false \\
 & DGCRN & 21.53 & 64.18 & 73.29 & 157.59 & \false & 13.28 & 43.64 & 14.36 & 206.60 & \false \\
 & UQGNN & 7.75 & 27.67 & 27.03 & 58.78 & \true & 3.47 & 11.98 & 11.53 & 22.99 & \true \\
 \midrule
\multirow{2}{*}{Attention} 
 & DSTAGNN & 15.67 & 41.84 & 48.09 & 99.82 & \true & 6.24 & 22.53 & 24.89 & 37.44 & \false \\
 & ASTGCN & 35.79 & 134.87 & 38.32 & 567.09 & \false & 12.64 & 45.84 & 11.44 & 202.17 & \true \\
\midrule
\multirow{2}{*}{Transformer} 
 & GluonTS & 12.84 & 33.58 & 66.15 & 81.06 & \false & 3.71 & 12.28 & 20.31 & 25.60 & \false \\
 & PatchTST & 9.39 & 27.50 & 45.27 & 60.62 & \false & 4.21 & 15.46 & 22.36 & 28.18 & \true \\
\midrule
\multirow{2}{*}{LLM} 
 & ST-LLM & 10.91 & 35.84 & 28.25 & 84.06 & \false & 4.21 & 15.77 & 12.25 & 28.90 & \false \\
 & UrbanGPT & 7.38 & 29.49 & 36.36 & 63.63 & \true & 3.69 & 13.40 & 15.83 & 23.55 & \true \\
\midrule
\multirow{2}{*}{Mamba} 
 & Mamba & \underline{7.17} & 27.18 & \underline{25.27} & \underline{58.53} & \true & \underline{3.16} & \textbf{10.73} & \underline{10.75} & 23.18 & \true \\
 & U-Mamba & 7.62 & \underline{25.88} & 28.47 & 59.72 & \true & 3.25 & 11.36 & 12.72 & \underline{22.69} & \false \\
\midrule
\rowcolor{gray!10}
Ours & \textbf{\m} & \textbf{6.78} & \textbf{25.24} & \textbf{23.38} & \textbf{56.82} & \true & \textbf{2.98} & \underline{10.89} & \textbf{10.69} & \textbf{22.24} & \true \\
\midrule\midrule

& & \multicolumn{5}{c}{\textbf{Texas}} & \multicolumn{5}{c}{\textbf{Florida}} \\
\cmidrule(lr){3-7} \cmidrule(lr){8-12}
\textbf{Category} & \textbf{Method} & MAE$\downarrow$ & RMSE$\downarrow$ & MPIW$\downarrow$ & IS$\downarrow$ & COV & MAE$\downarrow$ & RMSE$\downarrow$ & MPIW$\downarrow$ & IS$\downarrow$ & COV \\
\midrule
\multirow{5}{*}{GNN} 
 & DCRNN & 19.41 & 135.03 & 26.16 & 261.29 & \false & 140.31 & 317.69 & 305.84 & 648.91 & \false \\
 & STGCN & 16.79 & 124.16 & 57.43 & 125.01 & \false & 127.37 & 291.84 & 370.57 & 595.75 & \false \\
 & AGCRN & 6.54 & 41.32 & 16.59 & 70.71 & \false & 86.37 & 210.61 & 354.67 & 487.31 & \false \\
 & DGCRN & 24.41 & 152.62 & 41.73 & 414.77 & \false & 133.07 & 307.33 & 330.68 & 702.77 & \true \\
 & UQGNN & 5.40 & 37.38 & \textbf{17.87} & 46.08 & \true & 54.84 & 132.98 & \underline{193.39} & 352.42 & \false \\
 \midrule
\multirow{2}{*}{Attention} 
 & DSTAGNN & 10.46 & 68.95 & 37.84 & 72.46 & \false & 113.71 & 269.48 & 339.01 & 496.03 & \false \\
 & ASTGCN & 23.71 & 149.31 & 28.04 & 439.70 & \false & 129.01 & 316.88 & 271.01 & 659.53 & \false \\
\midrule
\multirow{2}{*}{Transformer} 
 & GluonTS & 5.69 & 38.24 & 28.39 & 46.24 & \false & 91.63 & 214.03 & 408.90 & 531.03 & \true \\
 & PatchTST & 21.90 & 145.26 & 19.07 & 341.24 & \true & 127.13 & 280.98 & \textbf{192.16} & 640.31 & \false \\
\midrule
\multirow{2}{*}{LLM} 
 & ST-LLM & 6.76 & 45.72 & 13.41 & 66.51 & \false & 75.37 & 170.38 & 258.19 & 467.20 & \true \\
 & UrbanGPT & 8.36 & 51.31 & 37.10 & 50.38 & \false & \underline{44.31} & \underline{119.76} & 232.98 & \underline{330.30} & \true \\
\midrule
\multirow{2}{*}{Mamba} 
 & Mamba & \underline{5.06} & \underline{35.84} & 18.64 & 46.59 & \true & 64.99 & 158.42 & 292.72 & 429.89 & \false \\
 & U-Mamba & 5.52 & 37.88 & 18.11 & \underline{45.63} & \true & 64.58 & 167.55 & 238.74 & 472.80 & \false \\
\midrule
\rowcolor{gray!10}
Ours & \textbf{\m} & \textbf{4.62} & \textbf{35.18} & \underline{18.15} & \textbf{44.50} & \true & \textbf{42.06} & \textbf{118.57} & 196.62 & \textbf{318.95} & \true \\
\bottomrule
\end{tabular}
\end{adjustbox}
\caption{Prediction performance for 1-day prediction horizon. $\downarrow$ indicates lower is better. The best results are in \textbf{bold} and the second-best are \underline{underlined}. \true~indicates target coverage ($\geq$90\%) achieved, while \false~indicates failure.}
\label{tab:horizon_1}
\end{table*}

\begin{table*}[tb]\small

\centering
\begin{adjustbox}{max width=\textwidth, keepaspectratio}
\begin{tabular}{@{}ll ccccc ccccc@{}}
\toprule
& & \multicolumn{5}{c}{\textbf{California}} & \multicolumn{5}{c}{\textbf{New York}} \\
\cmidrule(lr){3-7} \cmidrule(lr){8-12}
\textbf{Category} & \textbf{Method} & MAE$\downarrow$ & RMSE$\downarrow$ & MPIW$\downarrow$ & IS$\downarrow$ & COV & MAE$\downarrow$ & RMSE$\downarrow$ & MPIW$\downarrow$ & IS$\downarrow$ & COV \\
\midrule
\multirow{5}{*}{GNN} 
 & DCRNN & 36.85 & 120.24 & 90.32 & 284.13 & \false & 13.02 & 42.14 & 67.44 & 79.86 & \false \\
 & STGCN & 39.49 & 115.13 & 148.34 & 250.44 & \false & 10.86 & 41.24 & 45.13 & 67.22 & \false \\
 & AGCRN & 21.06 & 68.45 & 57.33 & 160.34 & \false & 7.82 & 26.49 & 21.28 & 57.81 & \false \\
 & DGCRN & 34.29 & 102.85 & 120.64 & 252.15 & \false & 21.25 & 69.85 & 23.64 & 331.79 & \false \\
 & UQGNN & 12.37 & 44.25 & 44.56 & 94.32 & \true & 5.57 & 19.17 & 19.90 & 37.89 & \false \\
 \midrule
\multirow{2}{*}{Attention} 
 & DSTAGNN & 24.95 & 67.18 & 79.25 & 160.42 & \false & 10.02 & 36.14 & 40.88 & 60.29 & \false \\
 & ASTGCN & 56.98 & 216.34 & 63.14 & 910.83 & \false & 20.32 & 73.58 & 18.85 & 325.64 & \false \\
\midrule
\multirow{2}{*}{Transformer} 
 & GluonTS & 20.47 & 53.82 & 108.99 & 130.23 & \false & 5.96 & 19.69 & 33.43 & 41.13 & \false \\
 & PatchTST & 14.98 & 44.16 & 74.51 & 97.39 & \false & 6.75 & 24.81 & 36.84 & 45.34 & \false \\
\midrule
\multirow{2}{*}{LLM} 
 & ST-LLM & 17.40 & 57.53 & 46.58 & 135.09 & \false & 6.76 & 25.29 & 20.16 & 46.54 & \false \\
 & UrbanGPT & 11.78 & 47.32 & 59.90 & 102.39 & \false & 5.91 & 21.49 & 26.06 & 37.90 & \true \\
\midrule
\multirow{2}{*}{Mamba} 
 & Mamba & \underline{11.44} & 43.62 & \underline{41.60} & \underline{94.05} & \false & \underline{5.07} & \textbf{17.21} & \underline{17.70} & 37.28 & \false \\
 & U-Mamba & 12.16 & \underline{41.57} & 46.90 & 95.95 & \false & 5.22 & 18.23 & 20.95 & \underline{36.43} & \false \\
\midrule
\rowcolor{gray!10}
Ours & \textbf{\m} & \textbf{10.84} & \textbf{40.71} & \textbf{38.53} & \textbf{91.24} & \true & \textbf{4.77} & \underline{17.48} & \textbf{17.59} & \textbf{35.71} & \true \\
\midrule\midrule

& & \multicolumn{5}{c}{\textbf{Texas}} & \multicolumn{5}{c}{\textbf{Florida}} \\
\cmidrule(lr){3-7} \cmidrule(lr){8-12}
\textbf{Category} & \textbf{Method} & MAE$\downarrow$ & RMSE$\downarrow$ & MPIW$\downarrow$ & IS$\downarrow$ & COV & MAE$\downarrow$ & RMSE$\downarrow$ & MPIW$\downarrow$ & IS$\downarrow$ & COV \\
\midrule
\multirow{5}{*}{GNN} 
 & DCRNN & 30.95 & 215.87 & 41.81 & 418.45 & \false & 224.07 & 508.24 & 488.72 & 1039.69 & \false \\
 & STGCN & 26.81 & 198.61 & 91.89 & 200.04 & \false & 203.62 & 467.41 & 592.82 & 954.15 & \false \\
 & AGCRN & 10.44 & 66.10 & 26.53 & 113.17 & \false & 138.16 & 337.53 & 567.45 & 780.18 & \false \\
 & DGCRN & 39.01 & 244.13 & 66.75 & 663.69 & \false & 212.90 & 492.10 & 529.22 & 1124.77 & \false \\
 & UQGNN & 8.64 & 59.83 & \textbf{28.58} & 73.76 & \true & 87.71 & 212.79 & \underline{309.20} & 564.03 & \false \\
 \midrule
\multirow{2}{*}{Attention} 
 & DSTAGNN & 16.72 & 110.33 & 60.54 & 115.93 & \false & 182.00 & 431.34 & 542.63 & 794.37 & \false \\
 & ASTGCN & 37.90 & 238.89 & 44.85 & 703.82 & \false & 206.36 & 507.16 & 433.78 & 1056.31 & \false \\
\midrule
\multirow{2}{*}{Transformer} 
 & GluonTS & 9.09 & 61.18 & 45.43 & 74.03 & \false & 146.60 & 342.70 & 654.24 & 850.40 & \false \\
 & PatchTST & 35.00 & 232.52 & 30.54 & 546.73 & \false & 203.49 & 449.79 & \textbf{307.46} & 1025.24 & \false \\
\midrule
\multirow{2}{*}{LLM} 
 & ST-LLM & 10.82 & 73.23 & 21.46 & 106.49 & \false & 120.65 & 272.64 & 412.89 & 747.83 & \false \\
 & UrbanGPT & 13.37 & 82.13 & 59.36 & 80.63 & \false & \underline{70.92} & \underline{191.71} & 372.97 & \underline{528.76} & \true \\
\midrule
\multirow{2}{*}{Mamba} 
 & Mamba & \underline{8.10} & \underline{57.42} & 29.82 & 74.58 & \true & 104.08 & 253.69 & 468.60 & 688.43 & \true \\
 & U-Mamba & 8.84 & 60.66 & 29.00 & \underline{73.05} & \false & 103.35 & 268.23 & 382.45 & 757.13 & \false \\
\midrule
\rowcolor{gray!10}
Ours & \textbf{\m} & \textbf{7.39} & \textbf{56.28} & \underline{29.05} & \textbf{71.38} & \true & \textbf{67.30} & \textbf{189.86} & 314.78 & \textbf{510.60} & \true \\
\bottomrule
\end{tabular}
\end{adjustbox}
\caption{Prediction performance for 7-day prediction horizon. $\downarrow$ indicates lower is better. The best results are in \textbf{bold} and the second-best are \underline{underlined}. \true~indicates target coverage ($\geq$90\%) achieved, while \false~indicates failure.}
\label{tab:horizon_2}
\end{table*}

\end{document}